\documentclass[11pt,a4paper]{article}
\usepackage[nohyperref]{naaclhlt2019}
\usepackage{times}
\usepackage{latexsym}
\usepackage{booktabs}
\usepackage{amsmath,amssymb}
\usepackage{multirow}
\usepackage{xcolor}
\usepackage{url}
\usepackage{graphicx}
\usepackage[utf8]{inputenc}
\usepackage{algorithm}
\usepackage{algorithmicx}
\usepackage[noend]{algpseudocode}

\aclfinalcopy % Uncomment this line for the final submission
\title{Improving Robustness of Machine Translation with Synthetic Noise}
\author{Vaibhav\thanks{\quad These authors contributed equally} , Sumeet Singh\footnotemark[1] , Craig Stewart\footnotemark[1] , Graham Neubig \\
  Language Technologies Institute \\
  School of Computer Science \\
  Carnegie Mellon University \\
  {\tt \{vvaibhav,sumeets,cas1, gneubig\}@cs.cmu.edu}}
  
\date{}

\begin{document}
\maketitle

\begin{abstract}
    Modern Machine Translation (MT) systems perform remarkably well on clean, in-domain text. However most human generated text, particularly in the realm of social media, is full of typos, slang, dialect, idiolect and other noise which can have a disastrous impact on the accuracy of MT. In this paper we propose methods to enhance the robustness of MT systems by emulating naturally occurring noise in otherwise clean data. Synthesizing noise in this manner we are ultimately able to make a vanilla MT system more resilient to naturally occurring noise, partially mitigating loss in accuracy resulting therefrom \footnote{Code available at \url{https://github.com/MysteryVaibhav/robust_mtnt}}.
\end{abstract}
\vspace{5pt}

\section{Introduction}

Machine Translation (MT) systems have been shown to exhibit severely degraded performance when required to translate of out-of-domain or noisy data \citep{luong15, sakaguchi16, belinkov17}. This is particularly pronounced when systems trained on clean, formalized parallel data such as Europarl \citep{europarl}, are tasked with translation of unedited, human generated text such as is common in domains such as social media, where accurate translation is becoming of widespread relevance \citep{mtnt}.

Improving the robustness of MT systems to naturally occurring noise presents an important and interesting task. Recent work on MT robustness \citep{belinkov17} has demonstrated the need to build or adapt systems that are resilient to such noise. We approach the problem of adapting to noisy data aiming to answer two primary research questions:
\begin{enumerate}
    \item Can we artificially synthesize the types of noise common to social media text in otherwise clean data?
    \item Are we able to improve the performance of vanilla MT systems on noisy data by leveraging artificially generated noise?
\end{enumerate}

In this work we present two primary methods of synthesizing natural noise, in accordance with the types of noise identified in prior work as naturally occurring in internet and social media based text \citep{eisenstein13, mtnt}. Specifically, we introduce a  \textbf{synthetic noise induction} model which heuristically introduces types of noise unique to social media text and \textbf{labeled back translation} \cite{bt}, a data-driven method to emulate target noise.

We present a series of experiments based on the Machine Translation of Noisy Text (MTNT) data set \citep{mtnt} through which we demonstrate improved resilience of a vanilla MT system by adaptation using artificially noised data.

\section{Related Work}

\citet{szegedy13} demonstrate the fragility of neural networks to noisy input. This fragility has been shown to extend to MT systems \citep{belinkov17,khayrallah2018impact} where both artificial and natural noise are shown to negatively affect performance. 

Human generated text on the internet and social media are a particularly rich source of natural noise \citep{eisenstein13, baldwin15} which causes pronounced problems for MT \citep{mtnt}. 

Robustness to noise in MT can be treated as a domain adaptation problem \citep{koehn17} and several attempts have been made to handle noise from this perspective. Notable approaches \citep{li10, axelrod11} include training on varying amounts of data from the target domain. \citet{luong15} suggest the use of fine-tuning on varying amounts of target domain data, and \citet{micelibarone17} note a logarithmic relationship between the amount of data used in fine-tuning and the relative success of MT models. 

Other approaches to domain adaptation include weighting of domains in the system objective function \citep{wang17} and specifically curated datasets for adaptation \citep{blodgett17}. \citet{kobus16} introduce a method of domain tagging to assist neural models in differentiating domains. Whilst the above approaches have shown success in specifically adapting across domains, we contend that adaptation to noise is a nuanced task and treating the problem as a simple domain adaptation task may fail to fully account for the varied types of noise that can occur in internet and social media text. 

Experiments that specifically handle noise include text normalization approaches \citep{baldwin15} and (most relevant to our work) the artificial induction of noise in otherwise clean data \citep{sperber17, belinkov17}.

\section{Data}

To date, work in the adaptation of MT to natural noise has been restricted by a lack of available parallel data. \citet{mtnt} recently introduced a new data set of noisy social media content and demonstrate the success of fine-tuning which we leverage in the current work. The dataset consists of naturally noisy data from social media sources in both English-French and English-Japanese pairs. 

In our experimentation we utilize the subset of the data for English to French which contains data scraped from Reddit\footnote{\url{www.reddit.com}}. The data set contains training, validation and test data. The training data is used in fine-tuning of our model as outlined below. All results are reported on the MTNT test set for French-English. We additionally use other datasets including Europarl (EP)~\cite{europarl} and TED talks (TED) \cite{ted} for training our models as described in \S \ref{exps}.

\begin{table}[h]
\centering{}
\small
\begin{tabular}{c c c}
  \toprule
 Training Data & \# Sentences & Pruned Size\\
  \midrule
	Europarl (EP) & 2,007,723 & 1,859,898\\ 
	Ted talk (TED) & 192,304 & 181,582\\
	Noisy Text (MTNT) & 19,161 & 18,112\\
  \bottomrule
\end{tabular}
\caption{Statistics about different datasets used in our experiments. We prune each dataset to retain sentences with length $\leq$ 50.}
\end{table}

\section{Baseline Model}

Our baseline MT model architecture consists of a bidirectional Long Short-Term Memory (LSTM) network encoder-decoder model with two layers. The hidden and embedding sizes are set to 256 and 512, respectively. We also employ weight-tying~\cite{presswolf} between the embedding layer and projection layer of the decoder. 

For expediency and convenience of experimentation we have chosen to deploy a smaller, faster variant of the model used in \citet{mtnt}, which allows us to provide comparative results across a variety of settings. Other model parameters reflect the implementation outlined in \citet{mtnt}.

In all experimental settings we employ Byte-Pair Encoding (BPE) \cite{bpe} using SentencePiece\footnote{\url{https://github.com/google/sentencepiece}}.

\begin{figure*}[t]
\begin{center}
\includegraphics[width=1.0\textwidth]{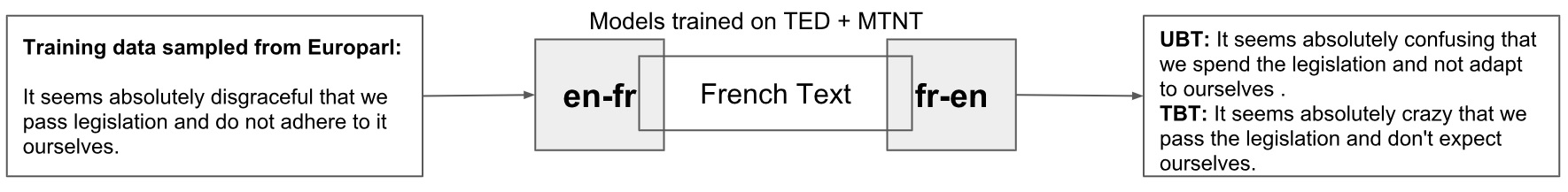}
\end{center}
\caption{Pipeline for injecting noise through back translation. For demostration purposes we show the process in an English sentence but in experiments, we use French sentences as input. % with reversed model order. %The baseline models trained for UBT, achieve a BLEU (on TED dev set) of 35.87 and 33.29 for fr-en and en-fr, respectively. Models trained for TBT acheive a BLEU (on TED + MTNT dev set) of 34.65 and 33.07 for fr-en and en-fr, respectively.
}
\label{fig:bt}
\end{figure*}
%\vv{TODO: Add details about using BPE for all our experiments and using multibleu script for computing BLEU.}
\section{Experimental Approaches \label{exps}}

We propose two primary approaches to increasing the resilience of our baseline model to the MTNT data, outlined as follows:

\subsection{Synthetic Noise Induction (SNI)}
For this method, we inject artificial noise in the clean data according to the distribution of types of noise in MTNT specified in \citet{mtnt}. For every token we choose to introduce the different types of noise with some probability on both French and English sides in 100k sentences of EP. % By setting the probability of these noise types according to the target domain, we expect to simulate its noise. 
%Figure \ref{fig:arti_noise} shows the impact of different types of noise and their level on BLEU score, computed on an otherwise clean newsdiscuss2015\footnote{\url{http://www.statmt.org/wmt15/test.tgz}} dev set.  
Specifically, we fix the probabilities of error types as follows: spelling (0.04), profanity (0.007), grammar (0.015) and emoticons (0.002). To simulate spelling errors, we randomly add or drop a character in a given word. %For spelling type we make a random choice to add or drop a character in the word. 
For grammar error and profanity, we randomly select and insert a stop word or an expletive and its translation on either side. Similarly for emoticons, we randomly select an emoticon and insert it on both sides.
Algorithm \ref{alg:arti_noise} elaborates on this procedure.

\begin{small}
\begin{algorithm}
\caption{Synthetic Noise Induction}
\label{alg:arti_noise}
\begin{algorithmic}
\State \textbf{Inputs:}{$\left[(p_1,\eta_1), (p_2,\eta_2) \cdots (p_k, \eta_k) \right]$}\Comment{{\small pairs of noise probabilities and noise functions}}
\Procedure{Add\_Noise}{$fr, en$}
\State $ o = 1 - \sum_i p_i $ \Comment{{\small probability of keeping original }}
\State $ D = [o, p_1, p_2, \cdots, p_k] $ \Comment{{ \small Discrete densities}}
\State $ j \gets \textsc{Select\_Index}(\textsc{Draw}(D))$ \Comment{{\small noise type}}
\If{$j \ne 0$} \Comment{{\small not original}}
   \State $ (fr,en) \gets \eta_j(fr,en) $ \Comment{{\small add noise to words}}
\EndIf
\State \textbf{return} $fr,en$
\EndProcedure
\end{algorithmic}
\end{algorithm}
\end{small}

\subsection{Noise Generation Through Back-Translation \label{bt-techs}}
We further propose two experimental methods to inject noise into clean data using the back-translation technique~\cite{bt}. %Figure \ref{fig:bt} elaborates the process.

\subsubsection{Un-tagged Back-Translation (UBT)}
We first train both our baseline model for fr-en and an en-fr model using TED and MTNT. % and also train a model for en-fr using the same data. 
%Once we have the forward and backward model, w
We subsequently take 100k French sentences from EP and generate a noisy version thereof by passing them sequentially through the trained models as shown in Figure \ref{fig:bt}. %Since the models are trained on different corpora 
The resulting translation will be inherently noisy as a result of imperfect translation of the intervening MT system.
%than the sentences being translated, we hope to obtain a noisy data which might be useful for training a noise resilient model as shown through our experiments.

\subsubsection{Tagged Back-Translation (TBT)}
The intuition behind this method is to generate noise in clean data whilst leveraging the particular style of the intermediate corpus.
%data on which we want to obtain superior performance, which is MTNT in our experiments. 
Both models are trained using TED and MTNT as in the preceding setting, save that we additionally append a tag in front on every sentence while training to indicate the origin data set of each sentence \cite{kobus16}. For generating the noisy version of 100k French sentences from EP, we append MTNT tag in front of the sentences before passing them through the pipeline shown in Figure \ref{fig:bt}.

%\begin{table*}[t]
%\centering{}
%\tiny
%\begin{tabular}{l l}
%  \toprule
%  & Output \\
%  \midrule
%	Input & \textit{That modernisation came to be particularly necessary for five reasons. Firstly, the authorisation system; secondly, decentralised application; thirdly,}\\&\textit{ procedural rules; fourthly, judicial application; and fifthly and finally, excessive red tape.}  \\
%	Un-tagged Back Translation & \textit{And this modernization has become particularly necessary for five reasons for the author of the author , at the time of the author application , at the}\\&\textit{ fourth way of the shallow application , and then to the very excessive red application .}  \\
%	Tagged Back Translation & \textit{This modernity was particularly necessary for five reasons, the system of engineering, a high decentralized application,}\\&\textit{ the rules of IT ... ... ... ... ... and finally, the red tape.}  \\
%  \bottomrule
%\end{tabular}
%\caption{Sample of obtained back translated Europarl data through the two techniques described in \S \ref{bt-techs}.}
%\label{ex3}
%\end{table*}

\section{Results}
We present quantitative results of our experiments in Table \ref{tbl:results}. Of specific note is the apparent correlation between the amount of in-domain training data and the resulting BLEU score.
%For each sub-table in Table \ref{tbl:results}, rows starting with + FT denotes fine-tuning the Baseline model mentioned in the first row of that particular sub-table. Our fine-tuning is nothing but continued training on existing model weights for 30 epochs or till convergence.
%We note that the amount of in-domain training data correlates with the BLEU score. 
The tagged back-translation technique produces the most pronounced increase in BLEU score of +6.07 points $(14.42 \longrightarrow 20.49)$.
%We also find that the tagged back translation technique of injecting noise in a clean data helps the most in making the model robust on a noisy dataset. The baseline model trained on just the Europarl data achieves a BLEU of 14.42 points on the MTNT test set without fine-tuning. If we fine-tune that baseline model with 100k TBT noisy Europarl data, the BLEU score jumps to 20.49 which is 6.07 points higher. 
This represents a particularly significant result given that we do not fine-tune the baseline model on in-domain data, attributing this gain to the quality of the noise generated.
%We argue that this is a strong result, because the gain in performance doesn't include direct use of in-domain data for fine-tuning the baseline model.  
%Instead, it is the quality of the noise addition in the clean data which leads to significant improvement. For other techniques of noise addition, we find that in those cases the performance without fine-tuning on in-domain data is similar to that of the baseline model.

\begin{table}[t]
\centering{}
\small
\begin{tabular}{c c c }
  %\selectlanguage{english}
  \toprule
  & Training data & BLEU \\
  \midrule
  \multicolumn{3}{c}{{\it Baselines}}\\
  \midrule
    Baseline & Europarl (EP) & 14.42 \\
	+ FT w/ & MTNT-train-10k & 22.49 \\
	+ FT w/ & MTNT-train-20k & 23.74 \\
	\midrule
	Baseline FT w/ & TED-100k & 10.92 \\
	+ FT w/ & MTNT-train-20k & 24.10 \\
  \midrule
  \multicolumn{3}{c}{{\it Synthetic Noise Induction}}\\
  \midrule
  Baseline FT w/ & EP-100k-SNI & 13.53\\
	+ FT w/ & MTNT-train-10k & 22.67 \\
	+ FT w/ & MTNT-train-20k & 25.05 \\
  \midrule
  \multicolumn{3}{c}{{\it Un-tagged Back Translation}}\\
  \midrule
  Baseline FT w/ & EP-100k-UBT & 18.71\\
	+ FT w/ & MTNT-train-10k & 22.75 \\
	+ FT w/ & MTNT-train-20k & 24.84 \\
  \midrule
  \multicolumn{3}{c}{{\it Tagged Back Translation}}\\
  \midrule
  Baseline FT w/ & EP-100k-TBT & 20.49\\
	+ FT w/ & MTNT-train-10k & \textbf{23.89} \\
	+ FT w/ & MTNT-train-20k & \textbf{25.75} \\
  \bottomrule
\end{tabular}
\caption{BLEU scores are reported on MTNT test set. MTNT valid set is used for fine-tuning in all the experiments. + FT denotes fine-tuning of the Baseline model of that particular sub-table, being continued training for 30 epochs or until convergence.
%Our fine-tuning is nothing but continued training on existing model weights for 30 epochs or till convergence.
}
\label{tbl:results}
\end{table}

\begin{table*}[t]
\centering{}
\scriptsize
% \fontsize{10}{12}\selectfont 
\begin{tabular}{l l}
  \toprule
  Systems & Output \\
  \midrule
	\textbf{REFERENCE} & $>$ And yes, I am an idiot with a telephone in usb-c... F*** that's annoying, I had to invest in new cables when I changed phones.  \\\hline
	\textbf{Baseline (trained on EP)} & And yes, I am an eelot with a phone in the factory ... P***** to do so, I have invested in new words when I have changed telephone.  \\\hline
	\textbf{FT w/ MTNT-train-20k} & $>$ And yes, I am an idiot with a phone in Ub-c. Sh**, it's annoying that, I have to invest in new cable when I changed a phone.  \\\hline
	\textbf{FT w/ EP-100k-TBT} & - And yes, I'm an idiot with a phone in the factory... Puard is annoying that, I have to invest in new cables when I changed phone.  \\\hline
	\textbf{FT w/ EP-100k-TBT} & $>$ And yes, I am an idiot with a phone in USb-c... Sh** is annoying that, I have to invest in new cables when I changed a phone.  \\\hspace{0.2cm}+ \textbf{MTNT-train-20k}&\\
  \bottomrule
\end{tabular}
\caption{Output comparison of decoded sentences across different models. Profane words are censored.}
\label{ex1}
\end{table*}

\begin{figure}
\centering{}
\includegraphics[width=0.5\textwidth]{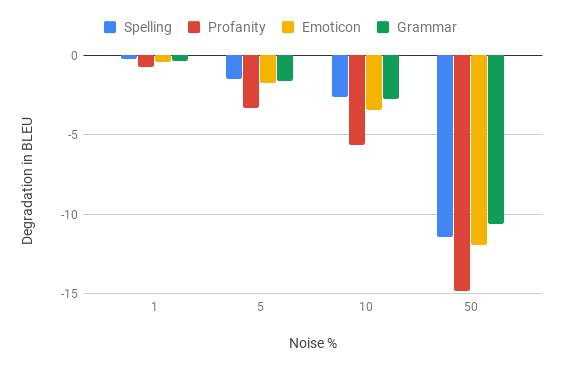}
\caption{The impact of varying the amount of Synthetic Noise Induction on BLEU.
%The impact of different kinds of noise addition on BLEU can be seen in the plot, as we make the dev set noisy. We use our Baseline model trained on EP for this study. Without any added noise in the newsdiscuss2015 dev set, the BLEU score achieved by the model is 17.56. This score drops as low as 2.74 when we introduce profanity noise with 50\% probability.
} 
\label{fig:arti_noise}
\end{figure}

The results for all our proposed experimental methods further imply that out-of-domain clean data can be leveraged to make the existing MT models robust on a noisy dataset. However, simply using clean data is not that beneficial as can be seen from the experiment involving \textit{FT Baseline w/ TED-100k}.

% \section{Analysis}
% In this section 

We further present analysis of both methods introduced above. Figure \ref{fig:arti_noise} illustrates the relative effect of varying the level of SNI on the BLEU score as evaluated on the newsdiscuss2015\footnote{\url{http://www.statmt.org/wmt15/test.tgz}} dev set, which is a clean dataset. From this we note that the relationship between the amount of noise and the effect on BLEU score appears to be linear. We also note that the most negative effect is obtained by including profanity. Our current approach involves inserting expletives, spelling and grammatical errors at random positions in a given sentence. However we note that our approach might under-represent the nuanced linguistic usage of expletives in natural text, which may result in its above-mentioned effect on accuracy. 

Table \ref{ex1} shows the decoded output produced by different models. We find that the output produced by our best model is reasonably successful at imitating the language and style of the reference. The output of \textit{Baseline + FT w/ EP-100k-TBT} is far superior than that of \textit{Baseline}, which highlights the quality of obtained back translated noisy EP through our tagging method. 

\begin{table}[t]
\centering{}
\scriptsize
\begin{tabular}{l l}
  \toprule
  Systems & Output \\
  \midrule
	\textbf{REFERENCE} & Voluntary or not because politicians are *very*\\&friendly with large businesses.  \\\hline 
	\textbf{FT w/ EP-100k-TBT} & Whether it's voluntarily, or invoiseally because\\&the fonts are *èsn* friends with the big companies.  \\\hline
	\textbf{FT w/ EP-100k-TBT} & Whether it's voluntarily, or invokes because the \\\hspace{0.2cm}+ \textbf{MTNT-train-10k}&politics are *rès* friends with big companies. \\\hline
	\textbf{FT w/ EP-100k-TBT} & Whether it's voluntarily, or invisible because the\\\hspace{0.2cm}+ \textbf{MTNT-train-20k}&politics are *very* friends with big companies. \\
  \bottomrule
\end{tabular}
\caption{Output comparison of decoded sentences for different amounts of supervision.}
\label{ex2}
\end{table}

We also consider the effect of varying the amount of supervision which is added for fine-tuning the model. From Table~\ref{ex2} we note that the \textit{Baseline + FT w/ EP-100k-TBT} model already produces a reasonable translation for the input sentence. However, if we further fine-tune the model using only 10k MTNT data, we note that the model still struggles with generation of *very*. This error dissipates if we use 20k MTNT data for fine-tuning. These represent small nuances which the model learns to capture with increasing supervision.

To better understand the performance difference between UBT and TBT, we evaluate the noised EP data.
%looked at the generated noisy version of English Europarl data. 
Figure \ref{fig:bt} shows an example where we can clearly see that the style of translation obtained from TBT is very informal as opposed to the output generated by UBT. Both the outputs are noisy and different from the input but since the TBT method enforces the style of MTNT, the resulting output is perceptibly closer in style to the MTNT equivalent.
%on the output to be MTNT styled, the back translation results are closer to actual MTNT data. 
This difference results in a gain of 0.9 BLEU of TBT over UBT.
%difference between the two methods.

\section{Conclusion}

This paper introduced two methods of improving the resilience of vanilla MT systems to noise occurring in internet and social media text: a method of emulating specific types of noise and the use of back-translation to create artificial noise. 
Both of these methods are shown to increase system accuracy when used in fine-tuning without the need for the training of a new system and for large amounts of naturally noisy parallel data.

\section{Acknowledgements}
The authors would like to thank the AWS Educate program for donating computational GPU resources used in this work.

\bibliographystyle{acl_natbib}
\bibliography{naaclhlt2019}

\end{document}